\documentclass[letterpaper, 10 pt, conference]{ieeeconf}

\IEEEoverridecommandlockouts

\usepackage{color}
\usepackage{multirow}
\usepackage{booktabs}
\usepackage{url}
\usepackage{subfigure}
\usepackage{threeparttable}
\usepackage{cite}
\usepackage{hyperref}[colorlinks,linkcolor=blue]

\renewcommand{\baselinestretch}{0.95}

\usepackage{enumerate}
\usepackage{amssymb}
\usepackage{leftidx}
\usepackage{amsfonts}
\usepackage{amsmath}
\usepackage{bm}
\usepackage{booktabs}
\usepackage{setspace}
\usepackage{makecell}
\usepackage{setspace}
\usepackage{float}
\usepackage{subfigure}
\usepackage[pdftex]{graphicx}
\graphicspath{{figs/}}
\usepackage{graphicx}
\graphicspath{{figs/}}
\newcommand{\etal}{\textit{et al}.}
\newcommand{\ie}{\textit{i}.\textit{e}.}
\newcommand{\eg}{\textit{e}.\textit{g}.}
\title{\LARGE \bf Robust Real-time LiDAR-inertial Initialization}
\author{Fangcheng Zhu$^{*}$, Yunfan Ren$^{*}$, Fu Zhang
\thanks{$^*$These two authors contribute equally to this work.}
\thanks{F. Zhu, Y. Ren and F. Zhang are with Department of Mechanical Engineering, University of Hong Kong. $\{ \textit{zhufc,renyf}\}$ $ \textit{@connect.hku.hk}$, $\textit{fuzhang@hku.hk}$}
}

\usepackage{tikz}
\usetikzlibrary{shapes,snakes}

\begin{document}
\thispagestyle{empty} 
\pagestyle{empty}  
\maketitle
\begin{tikzpicture}[overlay, remember picture]
  \path (current page.north) ++(0.0,-1.0) node[draw = black] {Accepted for the 2022 IEEE/RSJ International Conference on Intelligent Robots and Systems (IROS), Kyoto, Japan};
\end{tikzpicture}
\vspace{-0.3cm}

\pagestyle{empty}  
\thispagestyle{empty} 
\begin{abstract}
    For most LiDAR-inertial odometry, accurate initial states, including temporal offset and extrinsic transformation between LiDAR and 6-axis IMUs, play a significant role and are often considered as prerequisites. However, such information may not be always available in customized LiDAR-inertial systems. In this paper, we propose LI-Init: a full and real-time LiDAR-inertial system initialization process that calibrates the temporal offset and extrinsic parameter between LiDARs and IMUs, and also the gravity vector and IMU bias by aligning the state estimated from LiDAR measurements with that measured by IMU. We implement the proposed method as an initialization module, which can automatically detects the degree of excitation of the collected data and calibrate, on-the-fly, the temporal offset, extrinsic, gravity vector, and IMU bias, which are then used as high-quality initial state values for real-time LiDAR-inertial odometry systems. 
    Experiments conducted with different types of LiDARs and LiDAR-inertial combinations show the robustness, adaptability and efficiency of our initialization method. The implementation of our LiDAR-inertial initialization procedure LI-Init and test data are open-sourced on Github$\footnote{https://www.github.com/hku-mars/LiDAR\_IMU\_Init}$ and also integrated into a state-of-the-art LiDAR-inertial odometry system FAST-LIO2.
\end{abstract}


\section{Introduction}
Sensors are called the eyes of robots, which endow them with capability of exploring surroundings and performing self-localization and navigation. Camera is a commonly used sensor due to the ability to provide rich RGB information with low cost and light weight, but it is vulnerable to inadequate illumination and is lack of direct depth measurement, leading to high computation complexity when reconstructing 3D environments. Compared with cameras, light detection and ranging (LiDAR) sensors can offer direct, accurate 3D measurements and is robust to illumination changes, making it a preferred choice for robot localization\cite{xu2021fast,lin2020loam} and mapping \cite{liu2021balm} applications. 

To answer up the emergency such as sensor failure and to keep the whole system robust, multi-sensor fusion is becoming the main trend in recent years. Inertial Measurement Unit (IMU) is an excellent complementary sensor to fuse with camera or LiDAR since it can provide short-term ego-motion estimations without any external references. IMU is an ideal option to mitigate short-term odometry failure caused by degeneration, such as dim light scenes for cameras and structure-less environments for LiDARs. Moreover, high-frequency kinematic measurements from IMU help to compensate motion distortion of LiDAR scans especially when the robot is in high-speed motion\cite{ren2022bubble}. More and more multi-sensor based simultaneous localization and mapping (SLAM) methods show up, including visual-inertial system \cite{forster2016manifold, bloesch2017iterated}, LiDAR-inertial system \cite{xu2021fast, shan2020lio, liliom}, and LiDAR inertial visual system \cite{lin2021r2live, lin2021r3live}.

\begin{figure}[t]
    \setlength{\abovecaptionskip}{-0.1cm}
	\centering 
	\includegraphics[width=0.48\textwidth]{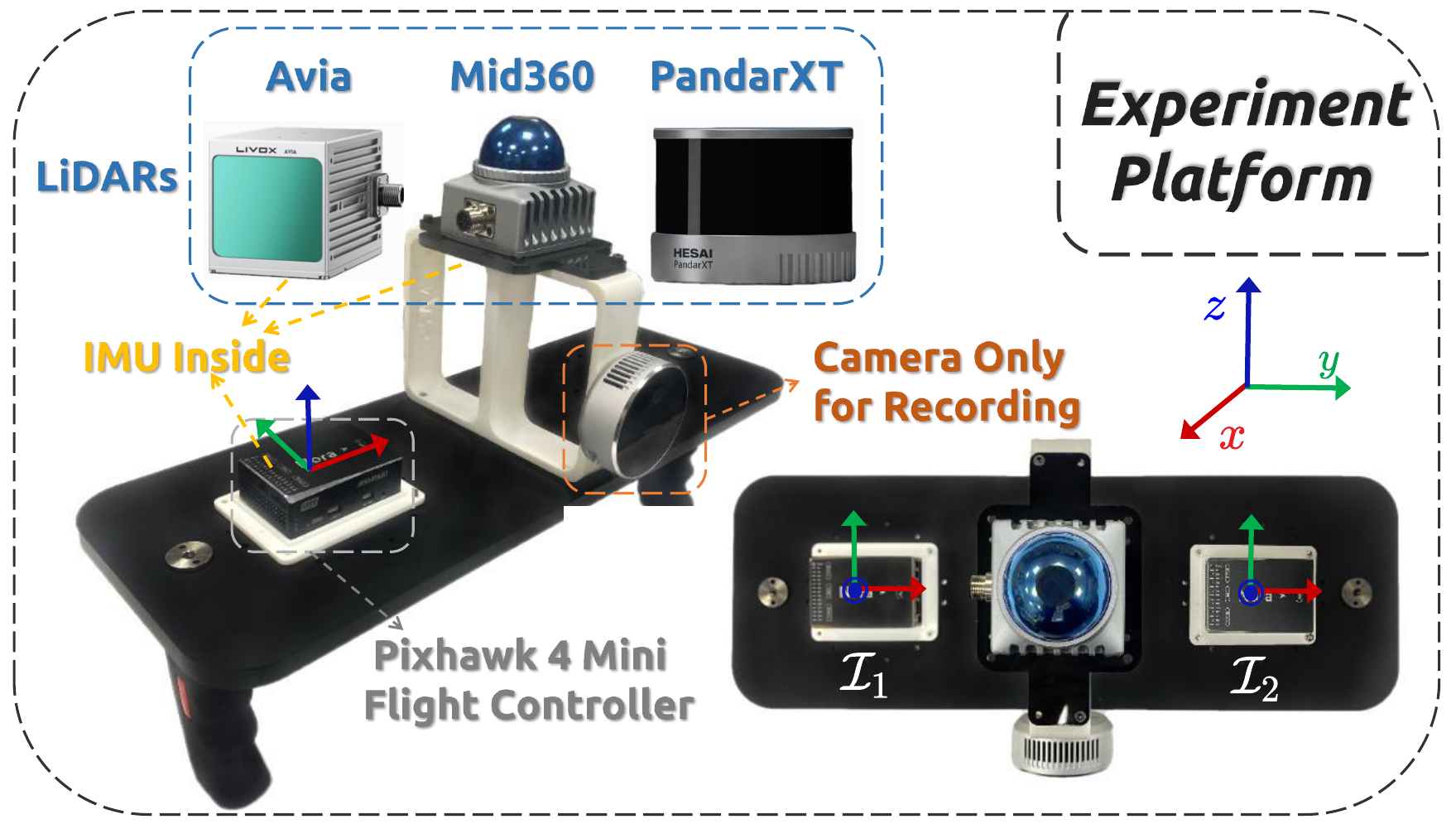}
	\caption{Experiment platform including multiple LiDARs (Non-repetitive scanning Livox Mid360, mechanical spinning Hesai PandarXT, small FoV Livox Avia) and built-in IMUs of LiDARs and Pixhawk flight controller. RealSense L515 camera is only used for recording videos in first personal view.}
	\label{fig:pic1}
\end{figure}

Owing to the strong non-linearity, the performance of sensor fusion system is heavily dependent on accurate initial states provided by efficient initialization module. Initialization of visual-inertial system has been widely studied \cite{mur2017visual,qin2017robust, huang2018online}, but few researches have focused on initialization for LiDAR-inertial system, which is necessary due to reasons below:  
1) For self-assembled devices, the LiDAR and IMU are often not time-synchronized and with unknown extrinsic, necessitating extra, laborious temporal and spatial calibration in advance.
2) Points of a LiDAR scan are sampled at different instants, leading to inevitable motion distortion. In case the temporal offset is unknown, IMU aided motion distortion compensation methods adopted by \cite{xu2021fast,lv2020targetless,mishra2021target} are no longer viable.
3) IMU raw measurements suffer from significant noises and the true values of linear accelerations and angular velocity are coupled with  unknown bias.
All these challenges drive us to find a well rounded LiDAR-inertial initialization method, capable of providing high-quality initial states including extrinsic transformation, gravity vector, IMU bias, and synchronizing the two sensors without any dedicated hardware setup.

Motivated by this, we propose a fast, robust LiDAR-inertial initialization method, which can automatically and accurately calibrate temporal offset and provide acceptable initial states without requiring any target or extra sensor, enabling a LiDAR-inertial odometry to run on a customized sensor setup without any dedicated prior calibration or hardware setup. Our contributions are highlighted as follows:

\begin{itemize}
\item We propose an efficient, accurate, hardware-free temporal calibration method based on cross-correlation and a unified temporal-spatial optimization, to estimate unknown but constant LiDAR-inertial temporal offset.
\item We propose a novel optimization formulation to perform spatial initialization and a  method to assess the degree of excitation in data. By further aligning states estimated from LiDAR with noise-mitigated IMU measurements, our initialization can automatically extract initialization data and estimate extrinsic transformation, gravity vector, gyroscope bias and accelerometer bias on the fly.
\item We conduct experiments on multiple types of LiDARs and LiDAR-inertial combinations (see Fig.~\ref{fig:pic1}) to validate the efficiency and accuracy of our initialization procedure. As far as we know, the proposed method is the first open-sourced temporal and spatial initialization algorithm for 3D LiDAR-inertial system, supporting both mechanical spinning LiDARs and non-repetitive scanning LiDARs.
\end{itemize}


\section{Related Works}
There is a wide variety of initialization methods for visual-inertial systems. For example,
an efficient IMU initialization method named VI-ORB-SLAM was introduced by Mur-Artal \etal \cite{mur2017visual}.
The initialization problem is divided into three simple sub-problems and achieves high accuracy in a short time.
Authors of \cite{qin2017robust} propose a robust initialization framework to recover the metric scale of monocular camera and to estimate extrinsic transformation and IMU bias.
Huang \etal \cite{huang2018online} propose a coarse-to-fine method to calibrate scale factor, gravity, and extrinsic transformation online.
For the calibration of temporal offset between camera and inertial sensors,
Mair \etal \cite{mair2011spatio} propose a method based on cross-correlation and phase congruency analysis, and calibrate extrinsic rotation following standard hand-eye calibration. Qin \etal \cite{qin2018online} propose an online method to calibrate temporal offset by jointly optimizing time offset, camera and IMU states.

In contrast to visual-inertial initialization, the initialization of LiDAR-inertial system is much less studied. Some of the existing initialization methods of LiDAR-inertial system require the sensors to be priorly synchronized, or rely on extra sensor, or ignore some initial states. Specifically, Wang \etal \cite{wang2021online} present an online initialization method to estimate the temporal and spatial offset between LiDAR and IMU, but camera is needed as an extra auxiliary sensor. Similarly, GPS/GNSS are used in \cite{taylor2015motion} for acquiring accurate position and attitude of the inertial sensor. Some LiDAR-inertial odometry systems have built-in initialization process, but these initialization modules are usually simple and incomplete. For example, \cite{xu2021fast} initializes the gyroscope bias, gravity vector, and temporal offset. But the initialization is fairly rough. For instance, the temporal offset is calibrated by assuming the sensor data receiving time as the sampling time, while data transmission and processing delay are totally neglected, leading to imprecise time offset estimation. The gyroscope bias in \cite{xu2021fast} is calibrated by keeping the sensor still for a few seconds in operation. Since gravity and accelerometer bias are coupled when the device stays still, accelerometer bias is not calibrated in its initialization. Although \cite{xu2021fast} calibrates the extrinsic online, the extrinsic initialization is not taken into account. So, good initial guess is required otherwise the convergence and robustness of the subsequent LiDAR-inertial odometry will be severely impacted. 
Similar to \cite{xu2021fast}, initialization of \cite{qin2020lins} requires the device to stay still for a while. The accelerometer bias and LiDAR-inertial extrinsic parameters are obtained by prior offline calibration while the temporal offset is assumed to be priorly known. Compared to \cite{wang2021online, taylor2015motion}, our proposed method does not require any extra sensor. Compared to \cite{xu2021fast, qin2020lins, fastlio1}, our work is more complete by initializing all the temporal offset, extrinsic, IMU bias, and gravity vector without any special requirements on the initial motion (\eg, keeping still) or any dedicated time synchronization or pre-calibration.

One of the main goals of our LiDAR-inertial initialization is to calibrate the extrinsic between LiDAR and IMU without any initial estimate. Some existing extrinsic calibration methods are based on batch optimization with tight data association, causing large time consumption. For example, Lv \etal \cite{lv2020targetless} propose a continuous-time batch optimization based calibration. The usage of B-splines leads to more parameters to be estimated and would result in large computation cost. \cite{mishra2021target} uses an extended Kalman filter to estimate the extrinsic transformation with complicated motion compensation, which has limited convergence speed. Compared with these methods, our method is more lightweight, being able to run on the fly, while still achieving accurate extrinsic calibration sufficient for subsequent online estimation (\eg, by \cite{xu2021fast}). Our methods also calibrates the temporal offset that are not considered in \cite{lv2020targetless,mishra2021target}. Besides, NDT based scan-to-scan matching adopted by \cite{lv2020targetless,mishra2021target} usually does not work well for LiDARs with non-repetitive scanning pattern. In contrast, our method adopt scan-to-map matching strategy, which can be easily applied to both repetitive and non-repetitive scanning LiDARs.

\section{Methodology}
\subsection{Framework Overview}
\begin{figure}[t]
    \setlength{\abovecaptionskip}{-0.2cm}
	\centering 
	\includegraphics[width=0.5\textwidth]{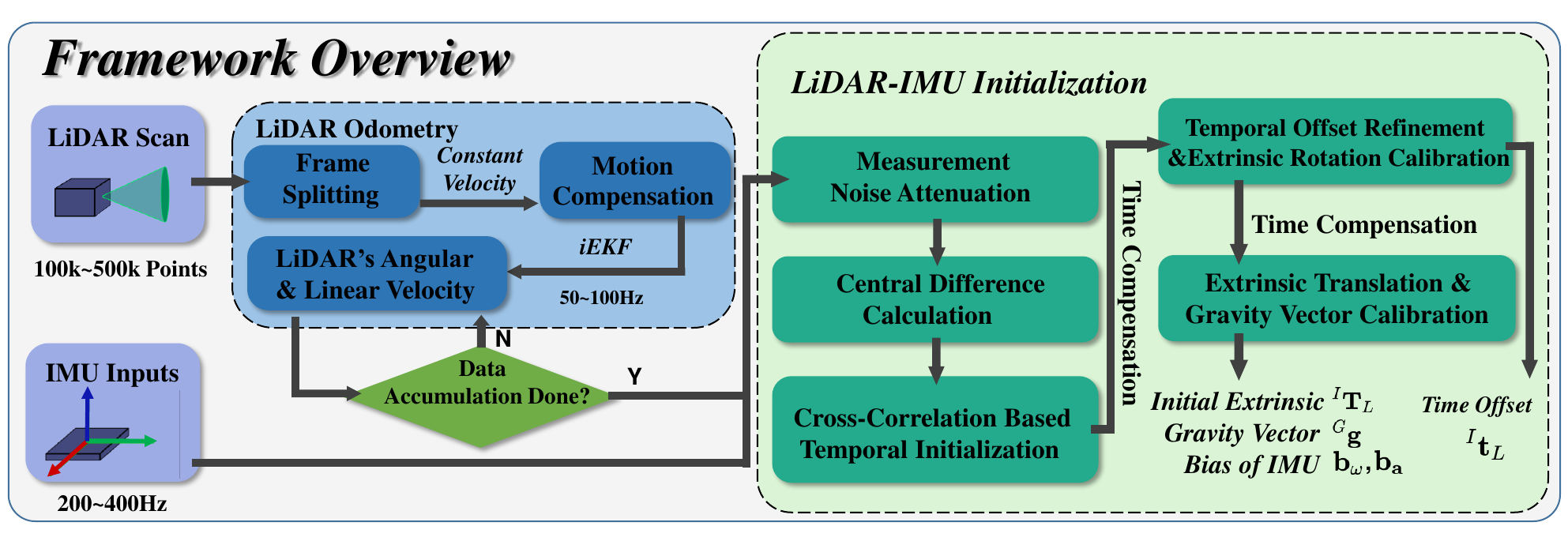}
	\caption{Framework of our LiDAR-inertial initialization procedure.}
	\label{fig:pic2}
\end{figure}

Since IMU is only excited when it is in motion \cite{mishra2021target}, our initialization procedure is a motion-based approach, which means sufficient excitation is necessary. The overview of our workflow is shown in Fig.~\ref{fig:pic2} and some important notations are shown in Table~\ref{notations}. The LiDAR odometry (see Section~\ref{section:Fast LO}) we propose is modified from FAST-LIO2 \cite{xu2021fast}, by adopting a constant (both angular and linear) velocity (CV) model to predict the LiDAR motion and compensate the point distortion in a scan. To mitigate the mismatch between the constant velocity model and the actual sensor motion, the LiDAR odometry rate is increased by splitting an input frame into several sub-frames. If the LiDAR odometry does not fail (\eg, due to degeneration) and the estimated LiDAR angular and linear velocity satisfy our proposed assessment criterion (see Section~\ref{section:assessment}), the excitation is considered to be sufficient and both LiDAR odometry output and the corresponding IMU data are fed to the initialization module (see Section.~\ref{section:LI Initialization}). In the initialization, the time offset is first calibrated by shifting IMU measurements to align with the LiDAR odometry, and then followed by an optimization process further refining the time offset, calibrating extrinsic transformation, and estimating IMU bias and gravity vector. The initialized states can be fed to a tightly-coupled LiDAR-inertial odometry (\eg, \cite{xu2021fast}) for online state estimation by fusing subsequent LiDAR and IMU data. 

\begin{table}[tbp]
	\renewcommand\arraystretch{1.2}
	\caption{Some Important Notations}
	\begin{center}
		\scalebox{0.78}{
			\begin{tabular}{l p{8.5cm} }
			\toprule
			\textbf{Notation} & \textbf{Explanation} \\ \hline
			$\boxplus/\boxminus$ &The encapsulated ``boxplus" and ``boxminus" operations on the state manifold. \\
			$t_k$ &Timestamp of the $k$-th LiDAR scan. \\ 
			$\rho_j$ &Timestamp of the $j$-th point in a LiDAR scan. \\
			$\tau_i$ &Timestamp of the $i$-th IMU measurement.  \\ 
			$L_j,L_k$ & The LiDAR body frame at the time $\rho_j$ and $t_k$.\\
			$\mathbf x,\widehat {\mathbf x}, \bar{\mathbf x}$ & The ground-true, predicted, and updated state value.\\
			$\Check{\mathbf x}$ & Estimation of $\mathbf x_j$ relative to $\mathbf x_k$ in backward propagation. \\
			${^I}\mathbf R_L,{^I}\mathbf p_L$ & The extrinsic rotation and translation from LiDAR to IMU.\\
			${^I} t_L$ & The total time offset between LiDAR and IMU.\\
			$\mathbf b_{\boldsymbol\omega}, \mathbf b_{\mathbf a}$ & The bias of gyroscope and accelerometer. \\
			${^G}\mathbf g$ & The gravity vector in global frame. \\
			$\mathcal I_i, \mathcal I_k$ & IMU data sequence used in initialization step with timestamp $\tau_i,t_k$ respectively. \\
			$\bar {\mathcal I}_k$ & IMU data sequence after compensating the initialized time offset, with synchronized timestamp $t_k$. \\
			$\mathcal L_k$ & LiDAR data used in initialization step with timestamp $t_k$. \\
			\bottomrule
		    \end{tabular}
		    }
	\end{center}
	\label{notations}
\end{table}

\subsection{LiDAR Odometry}\label{section:Fast LO}

Our LiDAR-only odometry and mapping is built on a constant velocity (CV) motion model, which assumes the angular and linear velocity are constant between two consecutive scans received at $t_k$ and $t_{k+1}$ respectively, \ie,
\begin{equation}\label{propagation}
    \mathbf x_{k+1} = \mathbf x_k \boxplus(\Delta t \mathbf f(\mathbf x_k,\mathbf w_k))
\end{equation}
where $\Delta t$ is the time interval between the two scans, the state vector $\mathbf x$, noise $\mathbf w$, and discrete state transition function $\mathbf f$ are defined as:
\begin{equation}
\mathbf x =
\begin{bmatrix}
^{G}\mathbf R_L\\
^{G}\mathbf p_L\\
^{G}\mathbf v_L\\
\boldsymbol \omega_L\\
\end{bmatrix},
\mathbf w=
\begin{bmatrix}
\mathbf n_{\mathbf v}\\
\mathbf n_{\boldsymbol \omega}
\end{bmatrix},
\mathbf f(\mathbf x,\mathbf w) =
\begin{bmatrix}
\boldsymbol \omega_{L}\\
^{G}\mathbf v_{L}\\
\mathbf n_{\mathbf v} \\
\mathbf n_{\boldsymbol \omega} \\
\end{bmatrix}
	\label{LO kinematic model}
\end{equation}
where $^{G}\mathbf R_L\in SO(3),{^G}\mathbf p_L$ are the attitude and position of LiDAR in the global frame (here is the first LiDAR body frame $L_0$), $^{G}\mathbf v_L$ is LiDAR's linear velocity described in global frame, and $\boldsymbol \omega_L$ is LiDAR's angular velocity in LiDAR body frame, which are modelled as a random walk process driven by Gaussian noises $\mathbf n_{\mathbf v}$ and $\mathbf n_{\boldsymbol \omega}$, respectively. In (\ref{propagation}), we used the notation $\boxplus/\boxminus$ defined in \cite{hertzberg2013integrating} to compactly represent the ``plus" on the state manifold. Specifically, for the state manifold $SO(3) \times \mathbb R^n$ in (\ref{LO kinematic model}), the $\boxplus$ operation and its inverse $\boxminus$ are defined as
\begin{equation*}
    \begin{bmatrix}
    \mathbf R \\ \mathbf a
    \end{bmatrix} \boxplus
    \begin{bmatrix}
    \mathbf r \\ \mathbf b
    \end{bmatrix}=
    \begin{bmatrix}
    \mathbf R\text{Exp}(\mathbf r) \\ \mathbf {a+b}
    \end{bmatrix};
    \begin{bmatrix}
    \mathbf R_1 \\ \mathbf a
    \end{bmatrix} \boxminus
    \begin{bmatrix}
    \mathbf R_2 \\ \mathbf b
    \end{bmatrix}=
    \begin{bmatrix}
    \text{Log}(\mathbf R_2^T \mathbf R_1) \\ \mathbf {a-b}
    \end{bmatrix}
\end{equation*}
where $\mathbf R, \mathbf R_1, \mathbf R_2 \in SO(3), \mathbf {r,a,b} \in \mathbb R^n$, $\text{Exp}(\cdot): \mathbb{R}^3 \mapsto SO(3) $ is the exponential map on $SO(3)$\cite{hertzberg2013integrating} and $\text{Log}(\cdot): SO(3) \mapsto \mathbb{R}^3$ is its inverse logarithmic map.

In practice, the sensor motion may not have a constant velocity. To mitigate the effect of this model error, we can split an input LiDAR scan into multiple sub-frames of smaller duration, over which the sensor motion agrees more with the CV model. 

\subsubsection{Error State Iterated Kalman Filter}\label{esikf}

Based on the on-manifold system representation (\ref{propagation}), we use an Error State Iterated Kalman Filter (ESIKF) \cite{ikfom} to estimate its states. The prediction step of the ESIKF consists of state prediction and covariance propagation as follows:
\begin{equation}\label{state_pred}
    \widehat {\mathbf x}_{k+1} = \bar {\mathbf x}_k \boxplus(\Delta t \mathbf f(\bar {\mathbf x}_k,\mathbf 0))
\end{equation}
\begin{equation}\label{cov propa}
    \widehat {\mathbf P}_{k+1} = \mathbf F_{\tilde {\mathbf x}}\bar {\mathbf P}_k \mathbf F_{\tilde {\mathbf x}}^T + \mathbf {F_w Q} \mathbf F_\mathbf w^T
\end{equation}
where $\mathbf {P,Q}$ are covariance matrix of state estimation and process noise $\mathbf w$ respectively. $\mathbf F_{\tilde {\mathbf x}}$ and $\mathbf {F_w}$ are as follows:
\begin{equation}
    \begin{aligned}
            \mathbf F_{\tilde {\mathbf x}} & \! = \!  
            \dfrac{\partial  \! \left(\!\left(\bar{\mathbf x}_k \! \boxplus \! \delta\mathbf x_k \! \right) \!\boxplus \! \left(\Delta t\mathbf f (\bar{\mathbf x}_k \! \boxplus \! \delta\mathbf x_k,\mathbf 0 ) \!\right) \! \boxminus \!
        ( \bar {\mathbf x}_k \! \boxplus \! \Delta t \mathbf f(\bar {\mathbf x}_k,\mathbf 0)\!)\! \right)}{\partial \delta \mathbf x_k} \\
            & = \begin{bmatrix}
            \text{Exp}(-\widehat {\boldsymbol \omega}_{L_k}\Delta t) & \mathbf 0_{3 \times 3} & \mathbf 0_{3 \times 3} & \mathbf I_{3 \times 3} \Delta t \\
            \mathbf 0_{3 \times 3} & \mathbf I_{3 \times 3} & \mathbf I_{3 \times 3} \Delta t & \mathbf 0_{3 \times 3} \\
            \mathbf 0_{3 \times 3} & \mathbf 0_{3 \times 3} & \mathbf I_{3 \times 3} & \mathbf 0_{3 \times 3}\\
            \mathbf 0_{3 \times 3} & \mathbf 0_{3 \times 3} & \mathbf 0_{3 \times 3} & \mathbf I_{3 \times 3}
         \end{bmatrix} \\
        \mathbf {F_w} &=  
        \dfrac{ \partial \left( \bar {\mathbf x}_k\boxplus \left( \Delta t \mathbf f( \bar{\mathbf x}_k ,\mathbf w_k) \right) \boxminus
        ( \bar {\mathbf x}_k\boxplus \Delta t \mathbf f(\bar {\mathbf x}_k,\mathbf 0)) \right)}{\partial \mathbf w_k} \\
        & = \begin{bmatrix}
            \mathbf 0_{3 \times 3} & \mathbf 0_{3 \times 3} \\
            \mathbf 0_{3 \times 3} & \mathbf 0_{3 \times 3} \\
            \mathbf I_{3 \times 3} \Delta t & \mathbf 0_{3 \times 3} \\
            \mathbf 0_{3 \times 3} & \mathbf I_{3 \times 3} \Delta t
         \end{bmatrix} 
    \end{aligned}
\end{equation}
where $\delta {\mathbf x}$ represents the error state.

\subsubsection{Motion Compensation}
In our considered problem, IMU and LiDAR are unsynchronized, hence IMU-aided motion compensation methods adopted by \cite{lv2020targetless,mishra2021target} are not viable. After receiving a new LiDAR scan at timestamp $t_{k+1}$, to compensate the motion distortion, we project each contained point $^{L_j}\mathbf p_j$ sampled at timestamp $\rho_j  \in (t_{k},t_{k+1})$ into the scan-end LiDAR frame $L_{k+1}$ as follows. 
With the constant velocity model, we have $^{G}\widehat {\mathbf v}_{L_{k+1}} = {^G} \bar{\mathbf v}_{L_{k}}$, $\widehat {\boldsymbol \omega}_{L_{k+1}} = \bar{\boldsymbol \omega}_{L_{k}}$, which leads to a relative transformation ${^{L_{k+1}}}\Check{\mathbf T}_{L_j}=({^{L_{k+1}}}\Check{\mathbf R}_{L_j},{^{L_{k+1}}} \Check{\mathbf p}_{L_j})$ from time $\rho_j$ to $t_{k+1}$ as:
\begin{equation}\label{motion distortion}
\begin{aligned}
{^{L_{k+1}}}\Check{\mathbf R}_{L_j} &= \text{Exp}(-\widehat {\boldsymbol\omega}_{L_{k+1}}\Delta t_j) \\
{^{L_{k+1}}}\Check{\mathbf p}_{L_j} &= - {^G}\widehat {\mathbf R}_{L_{k+1}}^T{^G}\widehat {\mathbf v}_{L_{k+1}}\Delta t_j\\
\Delta t_j &= t_{k+1} - \rho_j.
\end{aligned}
\end{equation}

Then the local measurement $^{L_j}\mathbf p_j$ can be projected to scan-end LiDAR frame as
\begin{equation}\label{project to tk}
     {^{L_{k+1}}}\mathbf p_j = {^{L_{k+1}}}\Check{\mathbf T}_{L_j} {^{L_j}}\mathbf p_j
\end{equation}

\begin{figure}[tbp]
    \setlength{\abovecaptionskip}{-0.15cm}
	\centering 
	\includegraphics[width=0.46\textwidth]{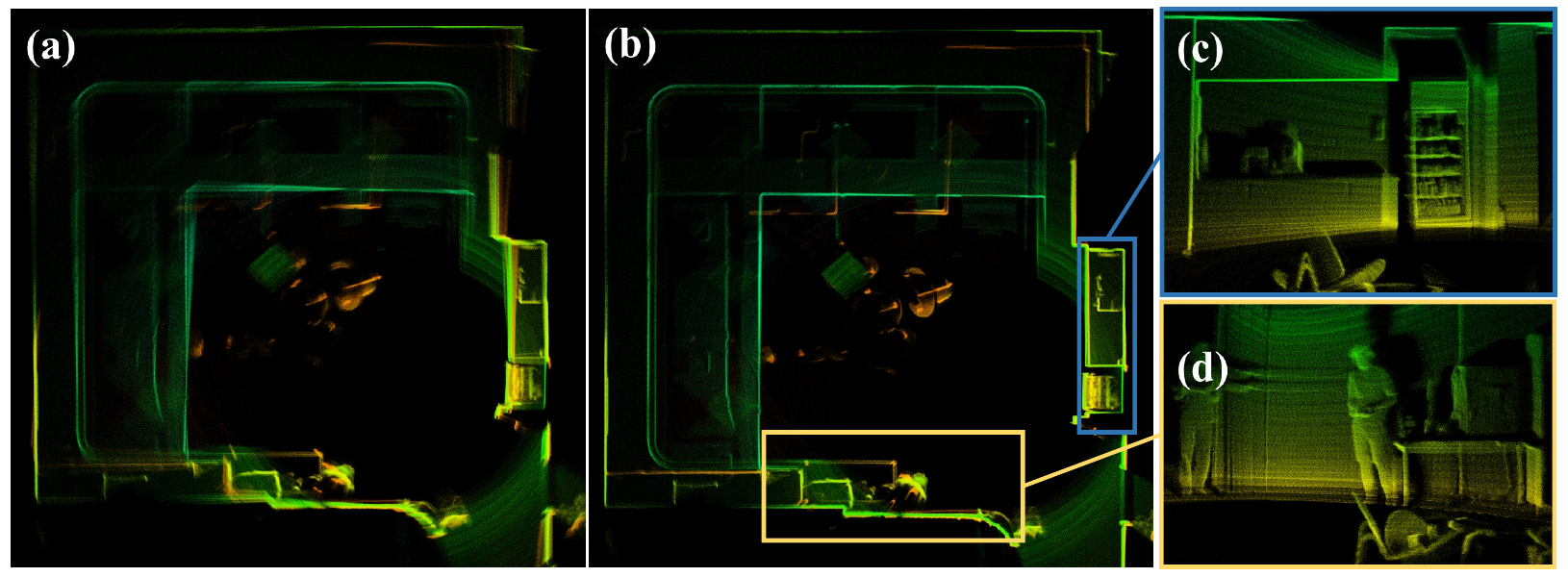}
	\caption{The mapping result comparison using Livox Mid360 LiDAR. (a) Map without motion compensation. (b) Point cloud map using distortion compensated scans following \eqref{motion distortion} and \eqref{project to tk}. (c, d) Mapping details of (b)}
	\label{fig:compensation}
\end{figure}

Then the distortion-compensated scan \{${^{L_{k+1}}}\mathbf p_j$\} provides an implicit measurement of the unknown state ${^G}\mathbf T_{L_{k+1}}$ expressed as the point-to-plane distance residual, based on which the full state $\mathbf x_{k+1}$ is iteratively estimated in an iterated Kalman filter framework until convergence. The converged state estimate, denoted as $\bar{\mathbf x}_{k+1}$, will then be used to propagate the subsequent IMU measurements as in Section \ref{esikf}. Details of this iterative estimation can be referred to FAST-LIO2 \cite{xu2021fast} or \cite{ikfom} for a more general treatment of manifold constraints. The mapping comparison of our LiDAR odometry using scans with and without motion compensation is shown in Fig. \ref{fig:compensation}.

\subsection{LiDAR-inertial Initialization}\label{section:LI Initialization}
The LiDAR odometry in Section \ref{section:Fast LO} outputs the LiDAR's angular velocity $\boldsymbol \omega_{L_k}$ and linear velocity $^{G}\mathbf v_{L_k}$ at each scan-end time with timestamp $t_k$. Meanwhile, IMU provides raw measurements, which are body angular velocity $\boldsymbol \omega_{m_i}$ and linear acceleration $\mathbf a_{m_i}$ with timestamp $\tau_i$. These data are accumulated and repeatedly assessed by the excitation criterion shown in Section.~\ref{section:assessment}. Once data of sufficient excitation is collected, the initialization module is called, which eventually outputs time offset ${^I} t_L \in \mathbb R$, extrinsic ${^I}\mathbf T_L = ({^I}\mathbf R_L,{^I}\mathbf p_L)\in SE(3)$, IMU bias $\mathbf b_\omega, \mathbf b_{\mathbf a}\in \mathbb R^3$, and gravity vector ${^G}\mathbf g \in \mathbb R^3$ in the global frame.

\subsubsection{Data Preprocess}
The IMU raw measurements suffer from noises $\mathbf n_{\omega_i}$ and $\mathbf n_{\mathbf a_i}$. The IMU measurement model is:
\begin{equation}
\begin{aligned}
        \boldsymbol \omega_{m_i} = \boldsymbol \omega_i^{\text{gt}} + \mathbf b_\omega + \mathbf n_{\omega_i}, \ \mathbf a_{m_i} = \mathbf a_i^{\text{gt}} + \mathbf b_{\mathbf a} + \mathbf n_{\mathbf a_i}
\end{aligned}
\end{equation}
where $\boldsymbol \omega_i^{\text{gt}}, \mathbf a_i^{\text{gt}}$ are the ground-truth of IMU angular velocity and linear acceleration. Similarly, the estimations $\boldsymbol \omega_{L_k}, {^{G}}\mathbf v_{L_k}$ from the LiDAR odometry contain noise as well.

To mitigate these noises, which are usually of high frequency, a non-causal zero phase low-pass filter\cite{gustafsson1996determining} is used to filter the noise without introducing any filter delay. The zero phase filter is implemented by running a Butterworth low-pass filter forward and backward \cite{gustafsson1996determining}, producing noise-attenuated IMU measurements $\boldsymbol \omega_{I_i} = \boldsymbol \omega_i^{\text{gt}} + \mathbf b_\omega, \mathbf a_{I_i} = \mathbf a_i^{\text{gt}} + \mathbf b_{\mathbf a}$. The noise-attenuated LiDAR estimations are still denoted as $\boldsymbol \omega_{L_k}, {^G}\mathbf v_{L_k}$ for notation simplicity.

From the LiDAR odometry $\boldsymbol \omega_{L_k}, {^G}\mathbf v_{L_k}$, we obtain the LiDAR angular and linear accelerations $\boldsymbol \Omega_{L_k}, {^G}\mathbf a_{L_k}$ by non-causal central difference \cite{centraldiff}. The resultant LiDAR odometry data can be collectively denoted as
\begin{equation}\label{eq:LiDAR-data}
\begin{aligned}
        \mathcal L_k &= \{\boldsymbol \omega_{L_k}, {^G}\mathbf v_{L_k}, \boldsymbol \Omega_{L_k}, {^G}\mathbf a_{L_k}\}
\end{aligned}
\end{equation}

Similarly, we obtain the IMU angular acceleration $\boldsymbol \Omega_{I_i}$ from noise-attenuated gyroscope measurements $\boldsymbol \omega_{I_i}$, leading to:
\begin{equation}
\begin{aligned}
        \mathcal I_i &= \{\boldsymbol \omega_{I_i}, \mathbf a_{I_i}, \boldsymbol \Omega_{I_i}\}
\end{aligned}
\end{equation}

\begin{figure}[h]
    \setlength{\abovecaptionskip}{-0.15cm}
	\centering 
	\includegraphics[width=0.45\textwidth]{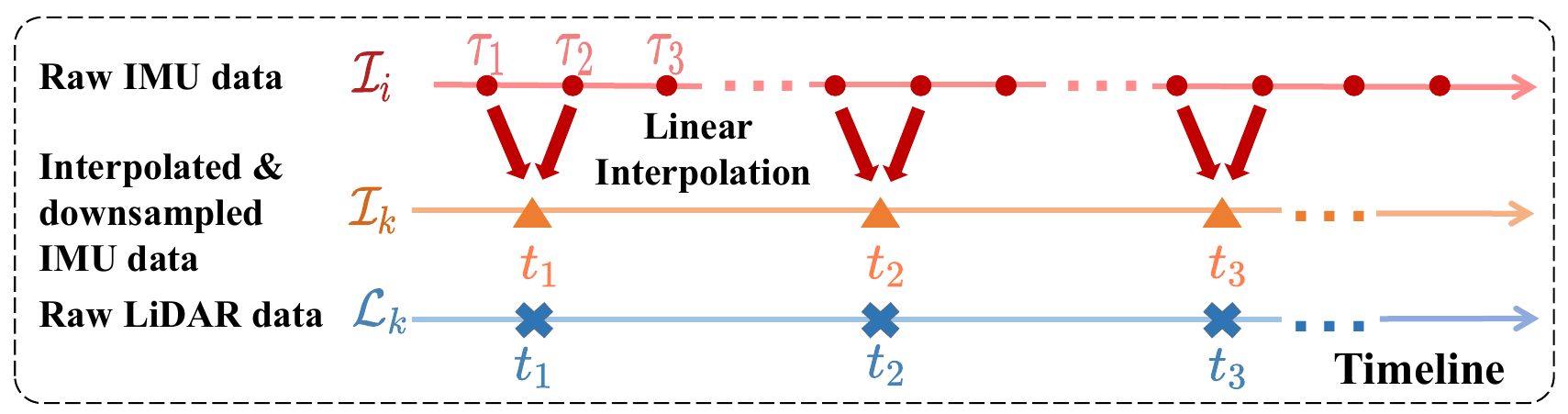}
	\caption{Down sample IMU data by interpolating it at each LiDAR odometry timestamp.}
	\label{fig:interpolation}
\end{figure}

Since IMU frequency is usually higher than that of LiDAR odometry, the two sequence $\mathcal I_i$ and $\mathcal L_k$ are not of the same size. To fix this, we extract the LiDAR and IMU data received within the same time period, and down-sample $\mathcal I_i$ by linearly interpolating it at each LiDAR odometry time $t_k$ (see Fig. \ref{fig:interpolation}). The down-sampled IMU data is  denoted as $\mathcal I_k$:

\begin{equation}\label{eq:IMU-data}
\begin{aligned}
        \mathcal I_k &= \{\boldsymbol \omega_{I_k}, \mathbf a_{I_k}, \boldsymbol \Omega_{I_k}\}
\end{aligned}
\end{equation}
which has the same timestamp $t_k$ with $\mathcal L_k$ (but the data is really delayed by the known temporal constant $^I t_L$). 

\subsubsection{Temporal Initialization by Cross-Correlation}
In most cases, due to inevitable transmission and processing delay prior to its reception by LiDAR-inertial odometry module, an unknown but constant offset ${^I} t_{L}$ between the LiDAR $\mathcal L_k$ and IMU $\mathcal I_k$ will exist, such that the IMU measurement $\mathcal I_k$, if advanced by ${^I} t_{L}$, will be aligned with the LiDAR odometry $\mathcal L_k$. Since the LiDAR data (\ref{eq:LiDAR-data}) and IMU data (\ref{eq:IMU-data}) are at discrete times $t_k$, advancing the IMU data is essentially made in discrete steps $d = {^I t_L}/{\Delta t}$, where $\Delta t$ is the time interval between two LiDAR scans. Specifically, for the angular velocity, we have
\begin{equation}\label{discrete_shift}
    \boldsymbol \omega_{I_{k+d}} = {^I}\mathbf R_L \boldsymbol \omega_{L_{k}} + \mathbf b_{\omega} 
\end{equation}

Ignoring the gyroscope bias $ \mathbf b_{\omega}$, which is usually small, we find that the magnitude of $\boldsymbol \omega_{I_{k+d}}$ and $\boldsymbol \omega_{L_{k}}$ should be the same, regardless of the extrinsic ${^I}\mathbf R_L$. Inspired by \cite{mair2011spatio}, we use the zero-centered cross-correlation to quantify the similarity between their magnitude. Then, the offset $d$ can be solved from the following optimization problem 
\begin{equation}\label{cross correlation}
   d^*  =  \mathop{\arg\max}_{d}  \sum    \|\boldsymbol \omega_{I_{k+d}}\|    \cdot  \|\boldsymbol \omega_{L_k}\|
\end{equation}
by enumerating the offset $d$ in the index range of $\mathcal L_k$.

\subsubsection{Unified Extrinsic Rotation and Temporal Calibration}\label{sec:rot_temp}
The cross-correlation method is robust against noise and small-scale gyroscope bias. But one obvious defect of \eqref{cross correlation} is that the calibration resolution of the temporal offset can only be made up to one sampling interval $\Delta t$ of the LiDAR odometry, any residual offset $\delta t$ smaller than $\Delta t$ cannot be identified. Let ${^I} t_{L}$ be the total offset between LiDAR odometry $\boldsymbol{\omega}_L$ and IMU data $\boldsymbol{\omega}_I$,  then ${^I} t_{L} = d^* \Delta t  + \delta t $. Similar to (\ref{discrete_shift}), the IMU measurement $\boldsymbol{\omega}_I$, if advanced by time ${^I} t_{L}$, will be aligned with the LiDAR odometry $\boldsymbol{\omega}_L$:
\begin{equation}\label{cont_shift}
    \boldsymbol \omega_{I}(t + {^I} t_{L}) = {^I}\mathbf R_L \boldsymbol \omega_{L}(t) + \mathbf b_{\omega} 
\end{equation}

Since the actual LiDAR odometry $ \boldsymbol \omega_{L}$ in (\ref{eq:LiDAR-data}) is only available at timestamps $t_k$, substituting $t=t_k$ and ${^I} t_{L} = d^* \Delta t  + \delta t $ into (\ref{cont_shift}) and noticing $\boldsymbol \omega_{L}(t_k) = \boldsymbol \omega_{L_k}$, we have
\begin{equation}\label{cont_shift1}
    \boldsymbol \omega_{I}(t_k + d^* \Delta t  + \delta t) = {^I}\mathbf R_L \boldsymbol \omega_{L_k} + \mathbf b_{\omega}. 
\end{equation}

\begin{figure}[t]
    \setlength{\abovecaptionskip}{-0.15cm}
	\centering 
	\includegraphics[width=0.48\textwidth]{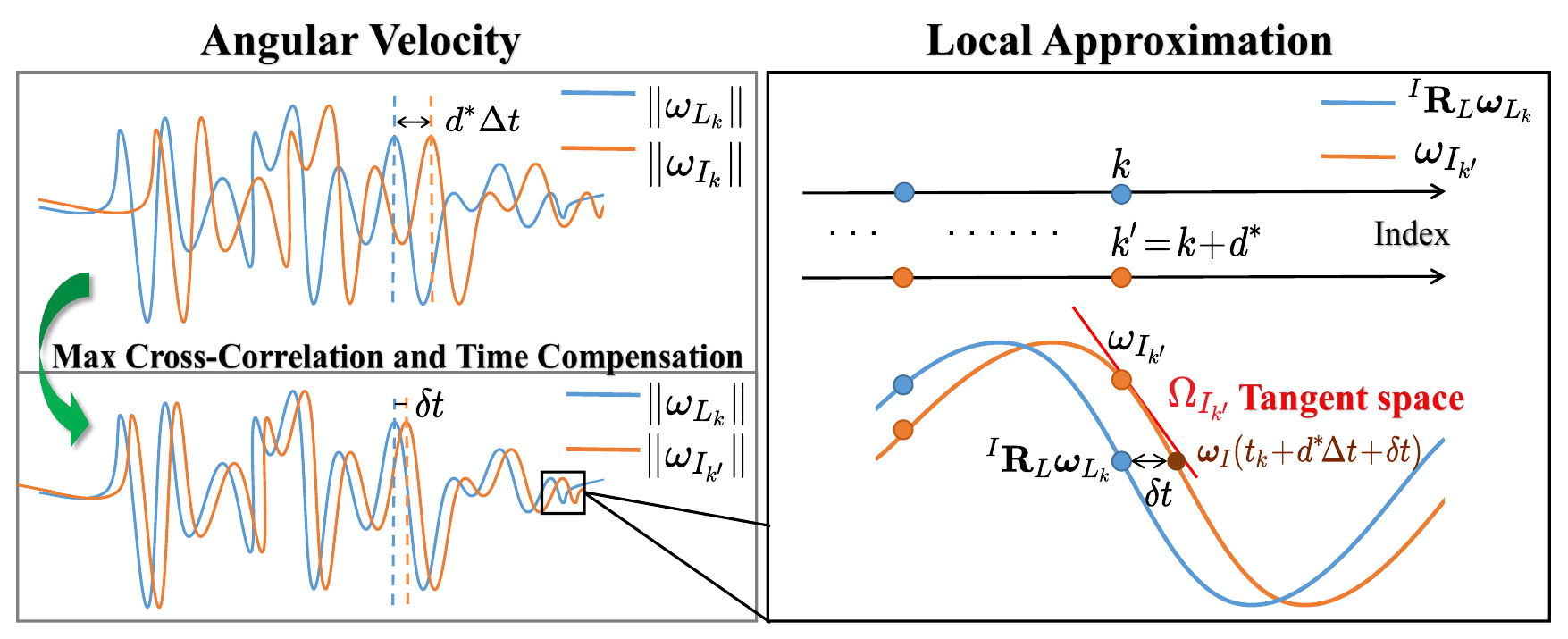}
	\caption{Illustration of time offset calibration and the first order approximation shown in Equation \eqref{omega_approx}.}
	\label{fig:pic4}
\end{figure}

Notice that $\boldsymbol \omega_{I}(t_k + d^* \Delta t  + \delta t)$ is the IMU angular velocity right after time $t_{k} +d^* \Delta t$, where the angular velocity and acceleration are $\boldsymbol \omega_{I}(t_k + d^* \Delta t )= \boldsymbol \omega_{I_{k'}}$ and $\boldsymbol \Omega_{I}(t_k + d^* \Delta t) = \boldsymbol \Omega_{I_{k'}}$, respectively, where $k' = k+d^*$. We can interpolate the value of $\boldsymbol \omega_{I}(t_k + d^* \Delta t  + \delta t)$ by assuming the angular acceleration is constant over the small $\delta t$ (see Fig. \ref{fig:pic4}):

\begin{equation}\label{omega_approx}
    \boldsymbol \omega_{I}(t_k + d^* \Delta t  + \delta t) \approx \boldsymbol \omega_{I_{k'}} + \delta t \boldsymbol \Omega_{I_{k'}}
\end{equation}
which can be substituted into (\ref{cont_shift1}) to obtain 
\begin{equation}\label{cont_shift2}
    \boldsymbol \omega_{I_{k'}} + \delta t \boldsymbol \Omega_{I_{k'}}  = {^I}\mathbf R_L \boldsymbol \omega_{L_k} + \mathbf b_{\omega} 
\end{equation}

Finally, based on the constraint in (\ref{cont_shift2}), the unified temporal-spatial optimization problem can be stated as:
\begin{equation}\label{unified calibration}
     \mathop{\arg\min}_{{^I}\mathbf R_L, \mathbf b_\omega, \delta t}\sum \|  {^I}\mathbf R_L \boldsymbol \omega_{L_k} \! + \! \mathbf b_\omega
         \! - \! \boldsymbol \omega_{I_{k'}} \! - \delta t\cdot \boldsymbol \Omega_{I_{k'}} \|^2
\end{equation}
which is solved iteratively (due to the nonlinear constraint ${^I}\mathbf R_L \in SO(3)$) by Ceres Solver$\footnote{http://ceres-solver.org/}$ from an initial value of $({^I}\mathbf R_L, \mathbf b_\omega, \delta t) = (\mathbf I_{3 \times 3}, \mathbf 0_{3 \times 1}, 0)$.

\subsubsection{Extrinsic Translation and Gravity Initialization}
In Section \ref{sec:rot_temp}, we obtained the extrinsic rotation ${^I}\mathbf R_L$, gyroscope bias $\mathbf b_\omega$ and the temporal offset $^{I}t_L$. In this section, we fix these values and proceed to the calibration of extrinsic translation, gravity vector, and acceleromter bias.

First, we align the IMU data $\mathcal I_k$ with that of LiDAR $\mathcal{L}_k$ using the offset $d^*$ and residual $\delta t$ previously calibrated. The aligned IMU data is denoted as $\bar{\mathcal I}_k$, which is now assumed to be perfectly aligned with $\mathcal{L}_k$ without time offset. Specifically, the IMU angular velocity $\bar{\boldsymbol \omega}_{I_k}$ corresponding to LiDAR angular velocity $\boldsymbol \omega_{L_k}$ at time $t_k$ is (see (\ref{cont_shift1}))
\begin{equation}
    \bar{\boldsymbol \omega}_{I_k} = \boldsymbol \omega_{I}(t_k + d^* \Delta t  + \delta t) \approx \boldsymbol \omega_{I_{k + d}} + \delta t \boldsymbol \Omega_{I_{k + d}}, 
\end{equation}

Similarly, the IMU acceleration $\bar{\mathbf a}_{I_k}$ corresponding to LiDAR acceleration $^G \mathbf a_{L_k}$ at time $t_k$ is
\begin{equation}
    \begin{aligned}
    \bar{\mathbf a}_{I_k} &= \mathbf a_{I}(t_k + d^* \Delta t  + \delta t) \\
    &\approx \mathbf a_{I_{k + d}} + \frac{\delta t}{\Delta t} (\mathbf a_{I_{k + d + 1}} - \mathbf a_{I_{k + d}} ), 
\end{aligned}
\end{equation}

Then, similar to \eqref{cont_shift}, we can find the acceleration constraint between IMU and LiDAR. As marked in \cite{xu2020robots}, the accelerations of two frames $A,B$ with fixed extrinsic have the following relationship: 
\begin{equation}\label{acc_relation}
    ^A\mathbf R_B\mathbf a_B=\mathbf a_A + \lfloor \boldsymbol \omega_A\rfloor_\wedge^2{^A}{\mathbf p}_B  + \lfloor\boldsymbol \Omega_A\rfloor_\wedge{^A}{\mathbf p}_B
\end{equation}
where $^A\mathbf R_B, ^A\mathbf p_B$ represent the extrinsic transformation from frame $B$ to frame $A$. Both $\mathbf a_A, \mathbf a_B$ are described in their own body frame.

For LiDAR-inertial system, we have two choices: $A$ for IMU and $B$ for LiDAR, or the opposite situation. Noticing that in the first case the accuracy of $ \boldsymbol \omega_A = \bar {\boldsymbol \omega}_{I_k} - \mathbf b_\omega$ is influenced by gyroscope bias estimation, and the error of $\boldsymbol \Omega_A$ would be amplified due to the noise in angular velocity measurements. To avoid this problem and increase the robustness of extrinsic translation calibration, we set LiDAR as $A$ and IMU as $B$. Since LiDAR's acceleration ${^G}\mathbf a_{L_k}$ is described in the global frame, we need to calculate LiDAR's instant acceleration described in body frame, denoted as $ \mathbf a_{L_k}$:
\begin{equation}\label{lidar_body_acc}
    \mathbf a_{L_k} = {^G}\mathbf R_L^T({^G}\mathbf a_{L_k} - {^G}\mathbf g)
\end{equation}
where ${^G}\mathbf R_L$ is the LiDAR's attitude in the global frame and is obtained from the LiDAR odometry in Section ~\ref{section:Fast LO}.

Finally, the extrinsic translation, accelerometer bias, and gravity vector can be jointly estimated from the following optimization problem:
\begin{equation}\label{trans calibration}
     \mathop{\arg\min}_{{^I}\mathbf p_L, \mathbf b_{\mathbf a}, {^G}\mathbf g}\sum \| {^I}\mathbf R_L^T(\bar{\mathbf a}_{I_k} \! - \! \mathbf {b_a}) 
    \! - \! \mathbf a_{L_k} \! - \! (\lfloor \! \boldsymbol\omega_{L_k} \! \rfloor_\wedge^2 \! + \! \lfloor \! \boldsymbol\Omega_{L_k} \! \rfloor_\wedge){^L}{\mathbf p}_I \|^2
\end{equation}
which can be solved iteratively (due to the constraint ${^G}\mathbf g \in \mathbb{S}_2$) by Ceres Solver from the initial value $({^I}\mathbf p_L, \mathbf b_{\mathbf a}, {^G}\mathbf g) = (\mathbf 0_{3\times1}, \mathbf 0_{3\times1}, 9.81\mathbf e_3) $. After ${^L}\mathbf p_I$ is estimated, the translation from LiDAR to IMU can be computed as ${^I}\mathbf p_L = -  {^I}\mathbf R_L{^L}\mathbf p_I$.

\subsubsection{Data Accumulation Assessment} \label{section:assessment}
The proposed initialization method relies on sufficient excitation (adequate motion) of LiDAR-inertial device. Thus, the system should be capable of assessing whether the excitation is sufficient to perform initialization all by itself. Ideally, the excitation can be assessed by the rank of the full Jacobian matrix of \eqref{unified calibration} for $({^I}\mathbf R_L, \mathbf b_\omega, \delta t)$ and \eqref{trans calibration} for $({^I}\mathbf p_L, \mathbf b_{\mathbf a}, {^G}\mathbf g)$. In practice, we found that it is sufficient to assess the Jacobian w.r.t. the extrinsic rotation ${^I}\mathbf R_L$ and extrinsic translation ${^I}\mathbf p_L$ only, since excitation on the extrinsic usually require complicated motion that excite the other states as well. Denote $\mathbf J_r$ the Jacobian of \eqref{unified calibration} w.r.t. ${^I}\mathbf R_L$ and  $\mathbf J_t$ the Jacobian of \eqref{trans calibration} w.r.t. ${^I}\mathbf p_L$, 
\begin{equation}
\begin{aligned}
        \mathbf J_r &= \begin{bmatrix}
        \vdots \\
        -{^I}\mathbf R_L  \lfloor \boldsymbol \omega_{L_k}\rfloor_\wedge \\
        \vdots
        \end{bmatrix}, 
    \mathbf J_t & = \begin{bmatrix}
    \vdots \\
    \lfloor \boldsymbol \omega_{L_k}\rfloor_\wedge^2 + \lfloor \boldsymbol \Omega_{L_k}\rfloor_\wedge \\
    \vdots
    \end{bmatrix}.
\end{aligned}
\end{equation}
Then the excitation can be assessed from the rank of $\mathbf J_r^T \mathbf J_r = \sum \lfloor \boldsymbol \omega_{L_k}\rfloor_\wedge^T \lfloor \boldsymbol \omega_{L_k}\rfloor_\wedge$ and $\mathbf J_t^T \mathbf J_t = \sum (\lfloor \boldsymbol \omega_{L_k}\rfloor_\wedge^2 + \lfloor \boldsymbol \Omega_{L_k}\rfloor_\wedge)^T(\lfloor \boldsymbol \omega_{L_k}\rfloor_\wedge^2 + \lfloor \boldsymbol \Omega_{L_k}\rfloor_\wedge)$. More quantitatively, the extent of excitation is indicated by the singular values of $\mathbf J_r^T \mathbf J_r$ and $\mathbf J_t^T \mathbf J_t$. Based on this principle, we developed an assessment program that can instruct the users how to move their devices to obtain sufficient excitation. We quantify the excitation based on the singular values of the Jacobian matrix, and set a threshold to assess if the excitation is sufficient.

\section{EXPERIMENTS}\label{experiments}
We evaluate our initialization method mainly on datasets collected by our self-assembled LiDAR-inertial handheld setup (Fig.~\ref{fig:pic1}). We test our initialization algorithm with multiple types of LiDAR (Livox$\footnote{https://www.livoxtech.com}$ Avia/Mid360 and Hesai PandarXT$\footnote{https://www.hesaitech.com/en/PandarXT}$) and 6-axis IMUs (Bosch BMI088 inside both Pixhawk flight controller$\footnote{https://ardupilot.org/copter/docs/common-cuav-nora-overview.html}$ and Livox LiDARs). The original data frequency of LiDAR is set as 10 $Hz$ (10 scans per second), the frequency of IMU raw data is 200 $Hz$. All the experiments are conducted on a desktop computer with Intel i7-10700 @2.90 $GHz$ with 32 $GB$ RAM. In all experiments, the initial states are set to $({^I}\mathbf R_L, \mathbf b_\omega, \delta t) = (\mathbf I_{3 \times 3}, \mathbf 0_{3 \times 1}, 0)$ in (\ref{unified calibration}) and $({^I}\mathbf p_L, \mathbf b_{\mathbf a}, {^G}\mathbf g) = (\mathbf 0_{3\times1}, \mathbf 0_{3\times1}, 9.81\mathbf e_3) $ in (\ref{trans calibration}). The initialization of other states does not need any initial values.

\subsection{Temporal Initialization Evaluation}
In order to show the effectiveness and accuracy of our temporal initialization (calibration) method, we test it on data collected by Livox LiDARs (Livox Avia and Livox Mid360) and their built-in IMUs. 
Since each Livox LiDAR and its built-in IMU are hardware synchronized in factory, the ground-true time offset is around 0 seconds, with microsecond accuracy level. To demonstrate the capability of our temporal calibration, we manually shift the input IMU timestamps to construct an artificial time offset \cite{qin2018online}, by adding a fixed value $ {^I} t_L$ to IMU timestamps. We collect 5 data sequences in a laboratory scene using each Livox LiDAR and its built-in IMU. The calibration results are shown in Table~\ref{temporal_calib_result}. As can be seen, the temporal calibration error is in microsecond level which suffices the requirements of most LiDAR-inertial sensor fusion algorithms. The results show our temporal calibration approach has great accuracy and consistency.
\begin{table}[tbp]
	\renewcommand\arraystretch{1.2}
	\caption{Temporal Initialization Results}
	\begin{center}
		\scalebox{0.78}{
			\begin{tabular}{p{1.2cm}<{\centering}|p{1.2cm}<{\centering}|p{2cm}<{\centering}|p{2cm}<{\centering} |p{1.5cm}<{\centering}}
			\toprule
			 \textbf{LiDAR} & ${^I} t_L$ (s) & \textbf{Mean}[s] & \textbf{RMSE}[s] &\textbf{NEES}\\ \hline
			\multirow{3} *{Avia} &0.05 & 0.0490 & 0.0016 &3.2$\%$ \\ \cline{2-5}
			&0.1 & 0.0988 & 0.0017 &1.7$\%$\\ \cline{2-5}
			&0.5 & 0.4989 & 0.0018 & 0.36$\%$ \\ \cline{1-5}
			\multirow{3} *{Mid360}  &0.05 & 0.0479 & 0.0034 & 6.8$\%$\\ \cline{2-5}
			&0.1 & 0.0982 & 0.0028 & 2.8$\%$ \\ \cline{2-5}
			&0.5 & 0.4984 & 0.0029 & 0.58$\%$ \\ \cline{2-5}
			\cline{1-4}
		    \end{tabular}
		}
	\end{center}
	\label{temporal_calib_result}
	\begin{tablenotes}
        \footnotesize
        \item
        Temporal initialization results of Livox LiDARs and their built-in IMUs. RMSE means root mean square error and NEES denotes normalized estimation error squared of calibrated time offset \cite{qin2018online}, computed as $\text{NEES}=\text{RMSE}/ {^I}t_L$.
    \end{tablenotes}
\end{table}

\begin{figure}[t]
    \setlength{\abovecaptionskip}{-0.4cm}
	\centering 
	\includegraphics[width=0.5\textwidth]{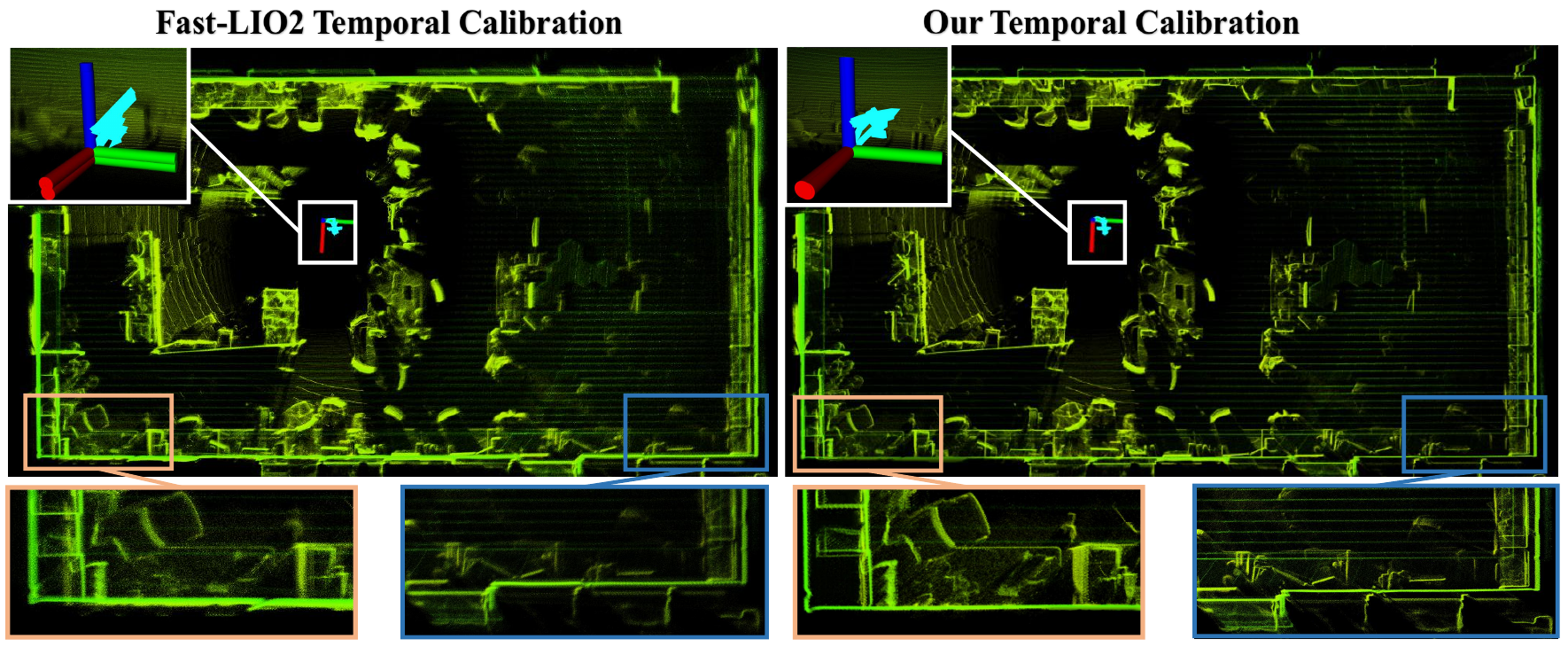}
	\caption{Mapping result comparison (bird view) using FAST-LIO2 on the same dataset collected in laboratory. The device is waved back and forth to obtain sufficient excitation, and return back to the origin. The entire sequence lasts 55.8 seconds, where the first 20 seconds are used to calibrate temporal offset. We rerun FAST-LIO2 on the entire sequence with the calibrated offset. (a) Mapping result with time synchronization in FAST-LIO2, the map accuracy is limited by the inaccurate internal synchronization.}
	\label{fig:pic5}
\end{figure}

For unsynchronized self-assembled LiDAR-inertial system, the ground-true time offset is hard to obtain. 
An indirect way that can validate the effectiveness of our approach is to examine the performance of tightly-coupled LiDAR-Inertial-Odometry (LIO) with the calibrated time offset. The LIO we adopt is FAST-LIO2 \cite{xu2021fast}, a state-of-the-art LiDAR inertial odometry that provides a temporal synchronization and state initialization internally. We test FAST-LIO2 with temporal offset calibrated by itself and by our approach, while keeping other state initialization unchanged with extrinsic from CAD reference, on data sequences collected by unsynchronized Hesai PandarXT LiDAR and Pixhawk IMU (Fig. \ref{fig:pic1}). The localization and mapping results are shown in Fig.~\ref{fig:pic5}. It can be seen that the overall LIO performance with our temporal calibration is better than that with the LIO's internal time synchronization. The odometry end-to-end drift is only 0.0102 m over a 11.3627 m trajectory for our method while 0.246 m for the LIO's internal time synchronization, the map with our approach also preserves more fine structural details.

\subsection{Spatial Initialization Evaluation}
\subsubsection{Multiple LiDAR Types Test}\label{multiple_lidar_test}
We test our spatial (extrinsic) initialization method on data sequences acquired by platform shown in Fig.~\ref{fig:pic1}. As the LiDARs are covered with shell and the IMU is built in Pixhawk flight controller, their precise extrinsic are hard to know. However, we can validate our extrinsic initialization approach in an indirect way. We fix the Pixhawk at two poses $\mathcal I_1$ and $\mathcal I_2$ shown in Fig.~\ref{fig:pic1}, their ground-true relative pose is designed in CAD and ensured in actual assembly. 
The ground-true relative pose is (0, 0, 0) degrees in Euler angle for rotation and (0.25, 0, 0) meters for translation, the manufacturing accuracy level of CAD can achieve 0.01 degrees and millimeters. In each pose, the extrinsic of LiDAR and IMU is calibrated, denoted as ${^{I_1}}\mathbf T_L$, ${^{I_2}}\mathbf T_L$, respectively. The relative pose is then computed as $^{I_1}\mathbf T_{I_2} = {^{I_1}}\mathbf T_L  {^{I_2}}\mathbf T_L^{-1}$, which can be compared with ground-truth provided by CAD reference. Aiming to prove that our calibration method supports multiple LiDARs with different scanning patterns, we collect 5 data sequences for each LiDAR, including Livox Mid360, Livox Avia, and Hesai PandarXT, with the Pixhawk IMU at both two poses. 

In all experiments, we set initial extrinsic as $({^I}\mathbf R_L, {^I}\mathbf p_L) = (\mathbf I_{3 \times 3}, \mathbf 0_{3 \times 1})$, even though the ground-truth is far from the initial values (\eg, for Mid360, IMU is at $\mathcal I_1$, the actual extrinsic is about $(0,-2, 178)$ degrees for rotation and $(0.12,0,0.11)$ meters for translation).
We calculate the absolute values of the relative IMU pose errors, in average and standard deviation. The results are shown in Table~\ref{extrinsic_calib_result}. As can be seen, the translation errors are in centimeter level, the rotational errors are less than 1$^{\circ}$. The overall small errors indicates the extrinsic initialization results are close across different datasets with the same sensor setup, and can be used as high-quality initial values for the subsequent LiDAR-inertial odometry.

\begin{table}[tbp]
	\renewcommand\arraystretch{1.4}
	\caption{Extrinsic Initialization Results}
	\begin{center}
		\scalebox{0.78}{
			\begin{tabular}{p{1.2cm}<{\centering}|p{1.2cm}<{\centering}|p{2.1cm}<{\centering}|p{2.1cm}<{\centering}|p{2.1cm}<{\centering}}
			\toprule
			 \multicolumn{2}{c|}{\textbf{LiDAR}} & Livox Mid360 & Livox Avia & PandarXT \\
			 \cline{1-5}
			\multirow{2}* {\makecell[c]{\textbf{Relative} \\ \textbf{Error}}  }
			& \textbf{Rot}($^{\circ}$) 
			& 0.2472$\pm$0.2043 & 0.4019$\pm$0.1708 & \textbf{0.7244}$\pm$\textbf{0.5076} \\
			\cline{2-5}
			& \textbf{Trans}(m)
			& 0.0081$\pm$0.0075 & 0.0064$\pm$0.0069 & \textbf{0.0133}$\pm$\textbf{0.0102} \\
			\cline{1-5}
		    \end{tabular}
		}
	\end{center}
	\label{extrinsic_calib_result}
	\begin{tablenotes}
        \footnotesize
        \item
        The mean value and SD (standard deviation) of extrinsic initialization. 
     \end{tablenotes}
\end{table}

Another interesting phenomenon from Table~\ref{extrinsic_calib_result} is that, the extrinsic calibration errors of Livox Avia and Livox Mid360 are smaller than that of Hesai PandarXT. This mainly benefits from the non-repetitive scanning of Livox LiDARs. Since our LiDAR odometry is based on constant velocity (CV) model, the input LiDAR frame is splitted into several subframes according to sampling time, to increase the odometry frequency and hence mitigate the CV model mismatch. For LiDARs with non-repetitive scanning, the frame splitting would not change the FOV of subframes. Also, the point map would get denser when the LiDAR stays still, which benefits our scan-to-map strategy. In contrast, for mechanical spinning LiDARs like Hesai PandarXT, the frame splitting would decrease the FOV of subframes and the robustness of LiDAR odometry. Thus, we can not move quickly when collecting data, which lowers the IMU excitation and the signal-to-noise ratio (SNR).

\begin{table}[tpb]
	\renewcommand\arraystretch{1.2}
	\caption{Extrinsic Calibration Comparison}
	\begin{center}
		\scalebox{0.78}{
			\begin{tabular}{c|c|c|c}
			\toprule
			\textbf{Method} &  \textbf{Rotation}($^{\circ}$) &\textbf{Translation}(m) & \textbf{Time}(s)\\\cline{1-4}
			\textbf{Proposed}& \textbf{0.6208} & \textbf{0.0162} & \textbf{10.2} \\ \cline{1-4}
			LI-Calib\cite{lv2020targetless} & 1.0375 & $\times$ &  332.6 \\ \cline{1-4}
			Target-Free\cite{mishra2021target} & 0.8483 & 0.0187 & 115.7 \\ \cline{1-4}
		    \end{tabular}
		}
	\end{center}
	\label{extrinsic_calib_compare}
	\begin{tablenotes}
        \footnotesize
        \item
        The first 40 seconds of the dataset is used for calibration.
        $\times$ denotes the refinement of LI-Calib\cite{lv2020targetless} fails and calibration result diverges due to ignorance of gravity vector initialization. So we use its rotation calibrated in the coarse calibration.
     \end{tablenotes}
\end{table}

 \begin{figure}[t]
    \setlength{\abovecaptionskip}{-0.15cm}
	\centering 
	\includegraphics[width=0.47\textwidth]{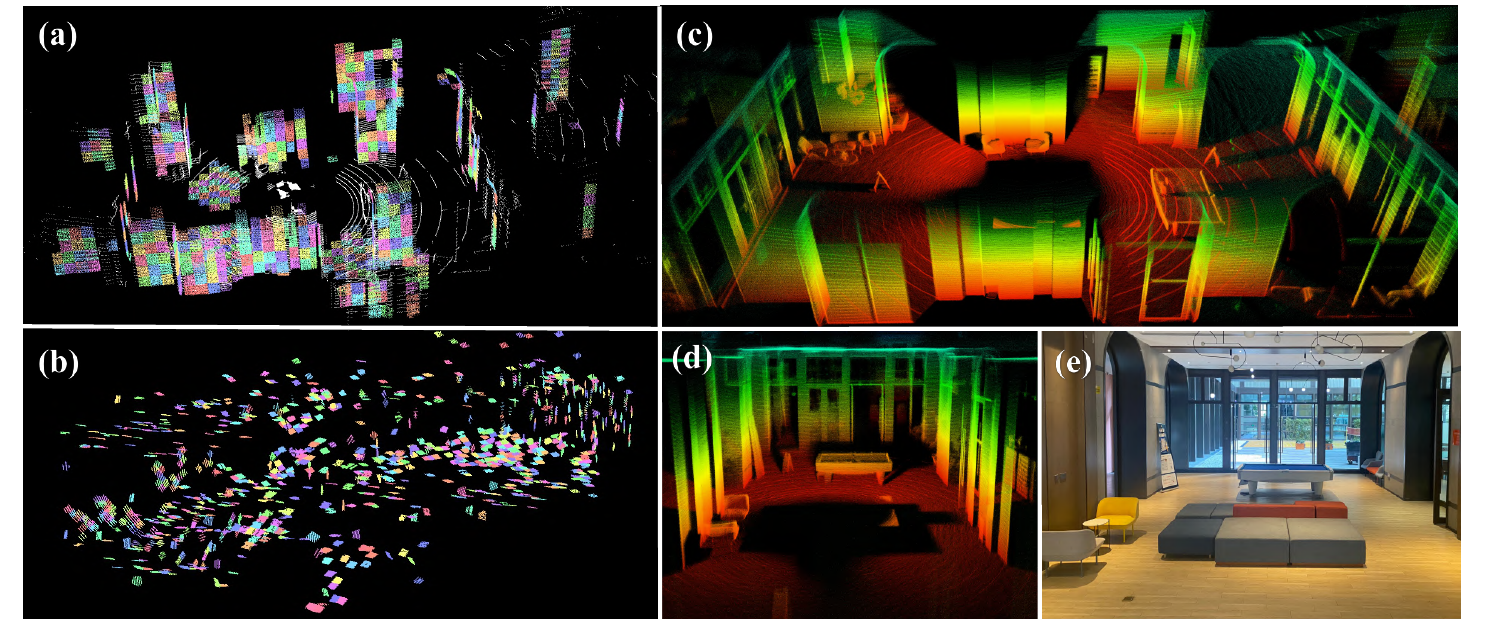}
	\caption{(a) Surfel map constructed during the coarse calibration of LI-Calib\cite{lv2020targetless}. (b) Refinement of LI-Calib fails, surfel map in a mess. (c,d) Accurate point cloud map constructed during our calibration, see Section ~\ref{section:Fast LO}. (e) The calibration scene.}
	\label{fig:pic6}
\end{figure}

\subsubsection{Accuracy and Robustness Comparison}
 To further verify the accuracy of our extrinsic initialization, we select the state-of-the-art LiDAR-inertial extrinsic calibration methods for comparison, which are LI-Calib \cite{lv2020targetless} and Target-Free \cite{mishra2021target}. Since these two methods only support mechanical spinning LiDAR, we select two sequences collected by Hesai PandarXT and Pixhawk in Section ~\ref{multiple_lidar_test}, and use the same relative IMU pose error to assess the calibration accuracy. Fig.~\ref{fig:pic6} (e) shows the scene of the two sequences. We use the first 40 seconds (including 400 LiDAR scans and corresponding IMU measurements) to run all calibration methods. The temporal offset is compensated in advance when testing LI-Calib and Target-Free since they are unable to do temporal calibration. The default parameters of the two methods are used on both sequences. The time-consumption and average relative IMU pose errors are shown in Table~\ref{extrinsic_calib_compare}. As can be seen, our method is more accurate while consuming much less time. Fig.~\ref{fig:pic6} shows some qualitative results. LI-Calib\cite{lv2020targetless} adopts a coarse-to-refine calibration. In the coarse calibration, it registers LiDAR scans by NDT matching and calibrates the extrinsic rotation; in the refine stage, it performs batch optimization of all the extrinsic parameters. The map constructed in the coarse calibration is shown in Fig.~\ref{fig:pic6} (a). The refinement of LI-Calib fails due to the ignorance of gravity initialization, leading to a messy surfel map shown in Fig.~\ref{fig:pic6} (b). In contrast, the LiDAR-only odometry in our calibration method is robust and accurate, the point map is shown in Fig.~\ref{fig:pic6} (c,d). Target-Free \cite{mishra2021target} does not provide visualization module to show its map. These results show our method has better robustness than \cite{lv2020targetless}, and similar accuracy compared with \cite{mishra2021target}, and much less computation cost than both \cite{lv2020targetless,mishra2021target}.

\begin{figure*}[tbp]
    \setlength{\abovecaptionskip}{-0.5cm}
	\centering 
	\includegraphics[width=1.0\textwidth]{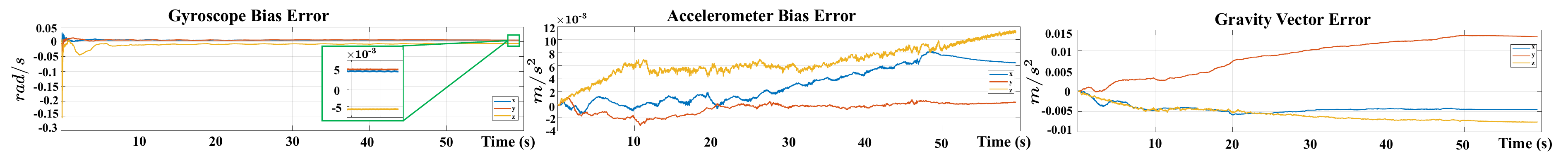}
	\caption{Error of gyroscope bias, accelerometer bias and gravity vector refined by FAST-LIO2 and their initial values supplied by our initialization. Test dataset is collected by Livox Mid360 LiDAR and IMU inside Pixhawk flight controller.}
	\label{fig:pic8}
\end{figure*}

\subsection{Time Consumption Evaluation}
Compared with \cite{lv2020targetless,mishra2021target}, two extrinsic calibration methods with high computation load, which cannot be processed in real-time, our approach is fast and can be implemented in real-time. Different from minimizing point-to-plane with batch optimization, our LiDAR odometry is very efficient, the average processing time of a subframe is about 8 ms. Once sufficient data is collected, the total time consumption of initialization solver, including data pre-processing, temporal initialization, extrinsic and gravity initialization is less than 500 ms, which is far smaller than the actual data collection time.

\begin{figure}[htbp]
    \setlength{\abovecaptionskip}{-0.15cm}
	\centering 
	\includegraphics[width=0.48\textwidth]{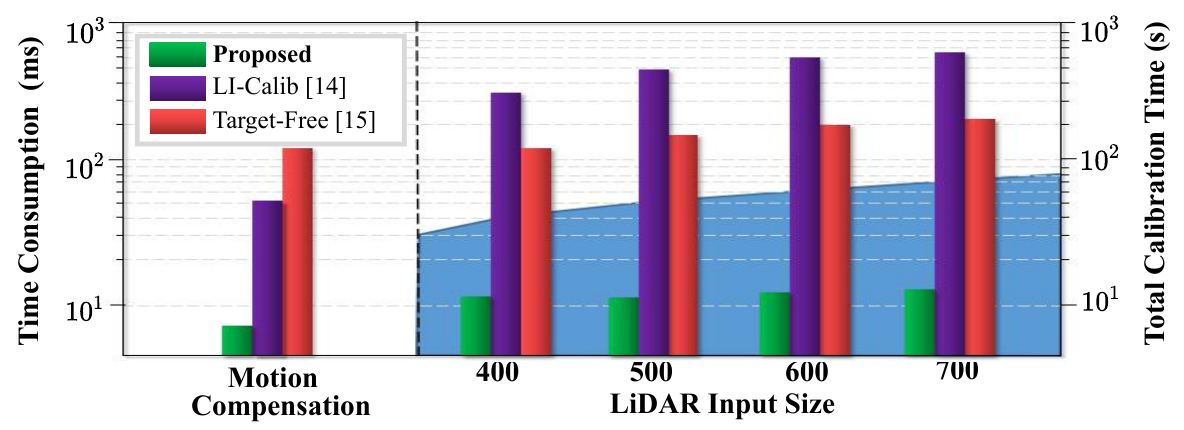}
	\caption{Time consumption comparison. All test data are collected by Hesai PandarXT LiDAR with 10 Hz output. The blue area is where the initialization time is below the data collection time, so it can run in real-time. }
	\label{fig:pic7}
\end{figure}

To evaluate the efficiency, we test \cite{lv2020targetless}, \cite{mishra2021target} and our method on the same dataset collected in an apartment hallway, and compare the time consumption of motion compensation for each scan. Also, we compare the total calibration time when the data length, measured by LiDAR input size, is different. All the results are shown in Fig.~\ref{fig:pic7}, which suggests that our method has high computation efficiency. Moreover, as the input data amount increases, the processing time of our approach grows slowly.

\subsection{Gravity and Bias Initialization Evaluation}
Our method is able to calibrate gyroscope bias, accelerometer bias and gravity vector, which can be used as high-quality initial states for real-time LiDAR-inertial odometry system. To examine the accuracy of our method, we integrate it into FAST-LIO2 \cite{xu2021fast}, which further refines all the states, including bias and gravity vector by tightly fusing the subsequent LiDAR data with IMU. We plot in Fig.~\ref{fig:pic8} the difference between the states estimated online by FAST-LIO2 and their initial values supplied by our initialization. As can be seen, initial gyroscope bias, accelerometer bias and gravity vector are already very accurate and the subsequent refinement are small.


\section{Conclusion}
This paper proposes a fast, robust, temporal and spatial initialization method for LiDAR-inertial system. An accurate and efficient coarse-to-fine temporal calibration method is proposed for unsynchronized LiDAR and IMU which is independent of any hardware synchronization setup. Also, we propose a fast, novel data association function to initialize LiDAR-inertial extrinsic transformation, gravity vector, as well as the bias of gyroscope and accelerometor. Various experiments show the consistency, robustness and high quality of our initialization method. Moreover, experiments using multiple types of LiDAR demonstrate the applicability to LiDARs with different scanning patterns.

\section{Acknowledgement}
This work is supported in part by the University Grants Committee of Hong Kong General Research Fund under Project 17206421 and in part by DJI under the grant 200009538. The authors gratefully acknowledge Livox Technology for the equipment support during the whole work. The authors would like to thank Wei Xu, Yixi Cai, Jiajun Lv, Meng Li for the helpful discussions and supports.

{\small
\bibliographystyle{unsrt}
\bibliography{egbib}
}

\end{document}